\documentclass{article}
\usepackage{ijcai25}

\usepackage{times}
\usepackage{soul}
\usepackage{url}
\usepackage[utf8]{inputenc}
\usepackage[small]{caption}
\usepackage{booktabs}
\usepackage[switch]{lineno}

\usepackage{amsmath,amssymb,amsfonts}
\usepackage{algorithmic}
\usepackage{graphicx}
\usepackage{textcomp}
\usepackage{xcolor}
\usepackage{multirow}
\usepackage{algorithm}
\usepackage{amsthm,amsmath,amssymb}
\usepackage{mathrsfs}
\usepackage{hyperref}
\usepackage{booktabs} 
\usepackage{siunitx}  

\def\BibTeX{{\rm B\kern-.05em{\sc i\kern-.025em b}\kern-.08em
    T\kern-.1667em\lower.7ex\hbox{E}\kern-.125emX}}
    
\hypersetup{colorlinks = true, 
	    linkcolor = black, 
	    urlcolor = blue,
           citecolor = blue} 

\title{Efficiently Training A Flat Neural Network Before It has been Quantizated}

\author{
Peng Xia
\and
Junbiao Pang$^1$
\and
Jiaxin Deng
\\\
\affiliations
$^1$Beijing University Of Technology\\
\emails
junbiao\_pang@bjut.edu.cn,
caitianyang@emails.bjut.edu.cn
}

\begin{document}

\maketitle

\begin{abstract}

Post-training quantization (PTQ) for vision transformers (ViTs) has garnered significant attention due to its efficiency in compressing models. However, existing methods typically overlook the relationship between a well-trained NN and the quantized model, leading to considerable quantization error for PTQ. However, it is unclear how to efficiently train a model-agnostic neural network which is tailored for a predefined precision low-bit model. In this paper, we firstly discover that a flat full precision neural network is crucial for low-bit quantization. To achieve this, we propose a framework that proactively pre-conditions the model by measuring and disentangling the error sources. Specifically, both the Activation Quantization Error (AQE) and the Weight Quantization Error (WQE) are statistically modeled as independent Gaussian noises. We study several noise injection optimization methods to obtain a flat minimum. Experimental results attest to the effectiveness of our approach. These results open novel pathways for obtaining low-bit PTQ models.

\end{abstract}

\section{Introduction}
\label{sec:intro}
Model compression has become an essential requirement for integrating deep models into edge computing devices. The prevalent methods in the domain of model compression include the search for optimal neural architectures~\cite{zoph-nas-arxiv-2016}, network pruning~\cite{han2-purn-arxiv-2015}, and the Deep Neural Network (DNN) quantization~\cite{li-brecq-arxiv-2021}~\cite{esser-lsq-arxiv-2019}. DNN quantization are categorized into two sub-classes: Post-Training Quantization (PTQ)~\cite{nagel-adaround-icml-2020},~\cite{li-brecq-arxiv-2021},~\cite{wei-qdrop-arxiv-2022},~\cite{li-ptqvit-tnnls-2023} and Quantization-Aware Training (QAT)~\cite{esser-lsq-arxiv-2019},~\cite{nagel-ooq-icml-2022}. PTQ adjusts the quantized model with a limited calibration dataset, bypassing the need for retraining. However, when dealing with low-bit widths, \textit{e.g.}, 2 or 4 bits, PTQ may face a significant drop in performance. 

Despite the demonstrated success of these advanced PTQ methods---for instance, QDrop~\cite{wei-qdrop-arxiv-2022} optimizes quantization policies by simulating information loss, while SmoothQuant~\cite{xiao2024smoothquantaccurateefficientposttraining} mitigates activation outliers in large models by transferring the quantization difficulty onto weights---we contend that they all operate under a fundamental constraint: their ultimate performance is capped by the intrinsic properties of the pre-trained, full-precision model.

Specifically, the central tenet of state-of-the-art PTQ approaches is to minimize a form of reconstruction error. This error quantifies the deviation of the quantized model's output from that of its full-precision counterpart, which is invariably treated as the ground truth. The objective of PTQ is, therefore, to emulate the functionality of the original model within the confines of a discrete parameter space. This framing, however, raises a critical and often-overlooked question: if the full-precision model itself is inherently sensitive to parameter perturbations, can any emulation of it be robust?

Empirical investigations reveal that high-performing full-precision models derived from standard training often converge to sharp loss landscapes. As illustrated in Figure~\ref{fig:trace}, these models exhibit acute sensitivity to minor parameter perturbations introduced by quantization. Even state-of-the-art PTQ algorithms merely attempt to ameliorate an inherently quantization-averse model, akin to balancing on a needle's point: regardless of the sophistication of the balancing technique, the intrinsic instability renders the endeavor precarious.

This instability arises from the deterministic perturbations imposed by quantization on model weights $\boldsymbol{W}$ and activations $\boldsymbol{x}$, denoted as $\Delta \boldsymbol{W}$ and $\Delta \boldsymbol{x}$. For a given input $\boldsymbol{x}$, the quantized model's loss $\boldsymbol{L_Q}$ can be approximated via Taylor expansion of the full-precision loss $\boldsymbol{L_{FP}}$:

\begin{equation}
\begin{split}
\boldsymbol{L_Q} & = \boldsymbol{L}(\boldsymbol{W} + \Delta \boldsymbol{W}, \boldsymbol{x} + \Delta \boldsymbol{x}) \\
 & \approx \boldsymbol{L}(\boldsymbol{W}, \boldsymbol{x}) + (\nabla \boldsymbol{L})^T \cdot [\Delta \boldsymbol{W}; \Delta \boldsymbol{x}] \\
& \quad + 0.5 \cdot [\Delta \boldsymbol{W}; \Delta \boldsymbol{x}]^T \cdot \boldsymbol{H} \cdot [\Delta \boldsymbol{W}; \Delta \boldsymbol{x}]
\end{split}
\end{equation}

where $\boldsymbol{H}$ denotes the Hessian matrix. This formulation elucidates that the loss degradation, $\boldsymbol{\Delta L} = \boldsymbol{L_Q} - \boldsymbol{L_{FP}}$, is predominantly governed by the second-order term, as the gradient $\nabla \boldsymbol{L}$ approaches zero at a loss minimum. Consequently, the norm of the Hessian $\|\boldsymbol{H}\|$, which quantifies the curvature or sharpness of the loss landscape, directly modulates the model's sensitivity to perturbations. A flat landscape (small $\|\boldsymbol{H}\|$) implies that quantization-induced perturbations $[\Delta \boldsymbol{W}; \Delta \boldsymbol{x}]$ yield minimal $\boldsymbol{\Delta L}$. Prevailing PTQ methods strive to minimize the perturbations $\Delta \boldsymbol{W}$ and $\Delta \boldsymbol{x}$ themselves through optimized quantization schemes; however, they overlook the potential to minimize $\|\boldsymbol{H}\|$ via tailored training processes.

In summary, contemporary PTQ paradigms lack a mechanism to proactively steer full-precision models toward minima that are inherently robust to quantization perturbations, i.e., flat minima.

Building upon this analysis, we introduce a novel paradigm termed Differential Noise-driven Quantization-aware Training \textbf{(DNQ)}, which implicitly regularizes the Hessian matrix to induce a flat loss landscape, thereby facilitating subsequent PTQ.

Intuitively, our approach eschews passive acceptance of a pretrained model; instead, it exposes the model to simulated quantization noise during training, compelling the optimizer to seek solutions that perform robustly not only at the current point but across its neighborhood, thereby naturally converging to flat minima. We explore this objective both theoretically and empirically, with the following key contributions:

\begin{itemize}
\item Theoretically, we reframe the challenge of achieving high-performance PTQ as optimizing the flatness of the full-precision model's loss landscape. We mathematically demonstrate the direct correlation between quantization-induced performance degradation and the Hessian norm, providing a rigorous foundation for training quantization-friendly models.
\item Methodologically, we propose the DNQ framework to implicitly minimize the Hessian norm. This framework periodically measures and models Weight Quantization Error (WQE) and Activation Quantization Error (AQE) via simulated PTQ. We innovatively employ differential noise injection for stable weight training, coupled with a two-stage strategy to balance convergence and robustness.
\item Empirically, our solution yields substantial advancements across multiple benchmark datasets and network architectures. Full-precision models trained with our solution, when subjected to simple PTQ, outperform those optimized with complex PTQ algorithms on standard pretrained baselines, validating the efficacy of our approach.
\end{itemize}

\label{fig:trace}

\section{Related Work}

\subsection{Post-Training Quantization}

Model quantization~\cite{DBLP:journals/corr/abs-1806-08342} is primarily pursued through Quantization-Aware Training (QAT)~\cite{esser-lsq-arxiv-2019}, ~\cite{nagel-ooq-icml-2022}, ~\cite{huang2024quantizationvariationnewperspective} and Post-Training Quantization (PTQ)~\cite{wei-qdrop-arxiv-2022}, ~\cite{ma-mrecg-cvpr-2023}, ~\cite{li-qdiffusion-cvpr-2023}, ~\cite{frantar2023gptqaccurateposttrainingquantization}. While QAT achieves high accuracy via resource-intensive retraining, PTQ has emerged as a lightweight alternative, converting a pre-trained model directly.

The PTQ methodology has evolved from minimizing layer-wise weight error to a data-driven, reconstruction-based paradigm. This modern approach, which underpins most SOTA methods, minimizes the feature map error between the FP and quantized models. This line of work was pioneered by AdaRound~\cite{nagel-adaround-icml-2020}, which used second-order information (Hessian) to guide layer-wise rounding. BRECQ~\cite{li-brecq-arxiv-2021} subsequently advanced this by extending the optimization granularity from layer-wise to block-wise, using the Fisher Information Matrix as a proxy for the Hessian to better compensate for inter-layer errors. This block-wise strategy is now a cornerstone in complex tasks, such as diffusion model quantization~\cite{li-qdiffusion-cvpr-2023}, ~\cite{sui-itsfusion-arxiv-2024b}.

Further refinements have focused on robustness. QDrop~\cite{wei-qdrop-arxiv-2022} links loss landscape flatness to generalization, introducing dropout during reconstruction to guide optimization towards flatter minima. MRECG~\cite{ma-mrecg-cvpr-2023} analyzes and mitigates error accumulation by adapting the reconstruction granularity.

Despite their sophistication, these PTQ methods are fundamentally post-hoc. They all operate on a given, pre-trained full-precision model. As we argued in Section~\ref{sec:intro}, if this initial model resides in a sharp, quantization-sensitive loss minimum, the efficacy of any post-hoc correction is inherently capped. These methods optimize the quantization process for a fixed landscape, whereas we argue that one must first optimize the landscape itself for quantization.

\subsection{Loss Landscape Shaping and Noise Injection}

To obtain a quantization-friendly model, it is crucial to find a solution in a wide, flat region of the loss landscape, as such solutions are more robust to perturbations. Stochastic Weight Averaging (SWA)~\cite{DBLP:journals/corr/abs-1803-05407} is a compelling solution, averaging weights from SGD iterates to locate the center of a wide optimal region. This concept is further illuminated from a Bayesian perspective by SWAG~\cite{DBLP:journals/corr/abs-1902-02476}, which connects the SGD trajectory to the geometry of the loss surface. Inspired by these findings, we incorporate SWA in the final stage of our framework.

Concurrently, the deliberate injection of stochastic noise is a long-standing technique for regularization and finding flatter minima. This is operationalized by perturbing parameters (e.g., PGD~\cite{jin2017escapesaddlepointsefficiently}), features (e.g., Dropout~\cite{wei-qdrop-arxiv-2022}, label smoothing~\cite{szegedy2015rethinkinginceptionarchitecturecomputer}), or by analyzing the implicit noise of the optimizer itself (e.g., SGD~\cite{mandt2018stochasticgradientdescentapproximate}, ~\cite{simsekli2019tailindexanalysisstochasticgradient}).

Recently, these noise-based principles have been adapted for quantization. However, their application remains distinct from our approach: QDrop~\cite{wei-qdrop-arxiv-2022} injects noise during the PTQ reconstruction phase, which is still a post-hoc operation. QAT methods~\cite{shin-nipq-cvpr-2023} inject pseudo-quantization noise during fine-tuning, which requires retraining.

While our work is inspired by this rich lineage, it is distinguished by a fundamental departure. Unlike the aforementioned methods, our framework introduces a principled methodology where the noise is statistically modeled to meticulously emulate the specific error distribution of the anticipated post-training quantization. Our core contribution is thus to re-purpose noise injection not as a generic regularizer, but as a targeted pre-conditioning tool designed to proactively forge a quantization-robust model before the PTQ process even begins.

\section{Proposed Method}

\textbf{Basic Notations.} In this paper, $\boldsymbol{x}$ represents a matrix (or tensor), a vector is denoted as  $\boldsymbol{x}$, $f(\boldsymbol{x};\boldsymbol{w}$) represents a FP model with the weight $\boldsymbol{w}$ and the input $\boldsymbol{x}$, $f(\boldsymbol{x};\boldsymbol{w}, \boldsymbol{s}, \boldsymbol{z})$ represents a quantized model with the parameter $\boldsymbol{w}$, quantization parameter $\boldsymbol{s}$, $\boldsymbol{z}$ and the input $\boldsymbol{x}$. We assume sample $\boldsymbol{x}$ is generated from the training set $\mathscr{D}_{t}$. 

\textbf{Quantization.} The channel-wise quantizer and layer-wise quantizer are adopted for weight and activation, respectively. For weights and the activation except for the post-Softmax activation, we adopt the uniform quantizer. Step size $\boldsymbol{s}$ and zero point $\boldsymbol{z}$ serve as a bridge between floating-point and fixed-point representations. Given the input tensor $\boldsymbol{x}$\footnote{It could either be feature $\boldsymbol{x}$ or weight $\boldsymbol{w}$.}, the uniform quantizer is defined as:
\begin{equation}\label{eqt:quantization}
    \begin{aligned}
    \boldsymbol{x}_{int} &= clip\left (   \lfloor{\frac{\boldsymbol{x}}{\boldsymbol{s}}}\rceil + \boldsymbol{z},0,2^{q}-1 \right ),\\
    \hat{\boldsymbol{x}} &=\left ( \boldsymbol{x}_{int}-\boldsymbol{z}  \right ) \boldsymbol{s},
    \end{aligned}
\end{equation}
where $\lfloor{\cdot  }\rceil$ represents the rounding-to-nearest operator, $q$ is the predefined quantization bit-width, $\boldsymbol{s}$ denotes the scale between two subsequent quantization levels.  $\boldsymbol{z}$ stands for the zero-points. Both $\boldsymbol{s}$ and $\boldsymbol{z}$ are initialized by a calibration set $\mathscr{D}_{c}$ from the training dataset $ \mathscr{D}_{t}$, \textit{i.e.}, $\mathscr{D}_{c}\in \mathscr{D}_{t}$.

\begin{eqnarray}
\boldsymbol{s} =\frac{\boldsymbol{x}_{max} - \boldsymbol{x}_{min}}{2^{q} - 1},\label{eqt:scale_init} \\
\boldsymbol{z} =\lfloor{q_{max}-\frac{\boldsymbol{x}_{max}}{\boldsymbol{s}}}\rceil,\label{eqt:zero_init}
\end{eqnarray}
where $q_{max}$ is the maximum value of the quantized integer.

\textbf{Objective.} Here, the quantization error induced by activation and weight quantization is denoted as $\delta \boldsymbol{x} = \hat{\boldsymbol{x}} - \boldsymbol{x}$ and $\delta \boldsymbol{W} = \overline{\boldsymbol{W}} - \boldsymbol{W}$. For each layer, we aim to minimize the Mean Squared Error (MSE) before and after weight-activation quantization: 
\begin{equation}\label{eqt:quantization_delta}
    \begin{aligned}
    \boldsymbol{L}^{MSE} &= \mathbb{E}[||\boldsymbol{W}\boldsymbol{x}-\overline{\boldsymbol{W}}\overline{\boldsymbol{x}}||^2_2] \\
    &=\mathbb{E}[||\boldsymbol{W}\boldsymbol{x}-\left(\boldsymbol{W} + \delta \boldsymbol{W} \right)\left(\boldsymbol{x} + \delta \boldsymbol{x} \right)||^2_2].
    \end{aligned}
\end{equation}
Eq.~\eqref{eqt:quantization_delta} indicates that output error is contributed both by activations and weight quantization error.

\subsection{Model the quantization error for both weight and activation}
\label{sec:method_overview}

A primary focus of our work is to train a model that is inherently robust to the quantization errors defined in Eq.~\eqref{eqt:quantization_delta}. However, directly minimizing this objective during training is intractable for several reasons. First, the quantization operator $\hat{(\cdot)}$ is non-differentiable. Second, the errors $\delta \boldsymbol{W}$ and $\delta \boldsymbol{x}$ are entangled and deterministic for any given model state, making it difficult to find a generalized solution that is robust to the perturbations encountered across the entire training trajectory.

To overcome these challenges, we propose a novel framework that reframes this deterministic optimization problem into a stochastic noise injection problem. Our key idea is to disentangle the weight and activation quantization errors and model them as independent, well-defined random variables. Specifically, we treat the quantization error for both weights and activations as samples drawn from a Gaussian distribution. This transformation from a deterministic error to a stochastic noise process allows us to leverage gradient-based optimization while forcing the model to adapt to a continuous space of perturbations, thereby implicitly finding a flat minimum in the loss landscape.

Our methodology is divided into two core components, which we term Weight Quantization Error Reduction (\textbf{WQER}) and Activation Quantization Error Reduction (\textbf{AQER}).

\begin{itemize}
    \item \textbf{WQER (Weight Quantization Error Reduction):} At the beginning of each training epoch, we perform a simulated per-channel PTQ on the current weights to measure the empirical weight quantization error (WQE), $\delta \boldsymbol{W}$. We then model this error distribution by estimating its per-channel mean and variance. These statistics are temporally smoothed using an Exponential Moving Average (EMA) to ensure stability.
    
    \item \textbf{AQER (Activation Quantization Error Reduction):} Similarly, we measure the activation quantization error (AQE), $\delta \boldsymbol{x}$, by running a calibration set through the current model. The per-tensor error distribution is then also modeled as a Gaussian, with its statistics similarly smoothed via EMA.
\end{itemize}

By statistically modeling WQE and AQE, we convert the intractable objective in Eq.~\eqref{eqt:quantization_delta} into a tractable problem of training a model to be robust against a specific, well-defined noise distribution. The subsequent sections will detail the precise mechanisms for injecting these modeled noises---a novel differential injection for weights and a stochastic drop-in for activations---to effectively and stably achieve this goal.

\subsection{Weight Quantization Error Reduction}
\label{sec:wqer}


\subsubsection{Statistical Modeling of WQE}
\label{sec:wqer_modeling}

At the beginning of each training epoch in the fine-tuning stage, we perform a simulated PTQ on the current full-precision weights $\boldsymbol{W}$ to obtain their quantized counterparts $\boldsymbol{W}_q$. The empirical WQE is then calculated as:
\begin{equation}\label{eqt:wqe_definition}
\boldsymbol{E}_w = \boldsymbol{W}_q - \boldsymbol{W} 
\end{equation}
Based on extensive empirical observation, we posit that the distribution of this error can be effectively approximated by a Gaussian distribution.

To capture the error characteristics accurately, we perform the statistical analysis at the same granularity as the quantization scheme itself. For convolutional layers, where per-channel quantization is standard, we compute the noise statistics for each output channel independently. Given a weight tensor of shape $[C_{out}, C_{in}, K_H, K_W]$, the error sub-tensor for the $i$-th output channel, $\boldsymbol{E}_{w,i}$, contains $N_i = C_{in} \cdot K_H \cdot K_W$ elements. The per-channel mean $\mu_{w,i}$ and variance $\sigma^2_{w,i}$ are thus computed as:
\begin{align}
    \mu_{w,i}   &= \frac{1}{N_i} \sum_{j,h,k} \boldsymbol{E}_{w,i,j,h,k} \label{eqt:conv_noise_mean} \\
    \sigma^2_{w,i} &= \frac{1}{N_i} \sum_{j,h,k} (\boldsymbol{E}_{w,i,j,h,k} - \mu_{w,i})^2 \label{eqt:conv_noise_var}
\end{align}
This yields a mean vector $\boldsymbol{\mu}_w \in \mathbb{R}^{C_{out}}$ and a variance vector $\boldsymbol{\sigma}^2_w \in \mathbb{R}^{C_{out}}$ for each convolutional layer. A similar per-channel (i.e., per-row) computation is performed for linear layers.

\subsubsection{Differential Noise Injection for Weights}
\label{sec:wqer_injection}

Having modeled the WQE distribution $P_w = \mathcal{N}(\boldsymbol{\mu}_w, \boldsymbol{\sigma}^2_w)$ in Section~\ref{sec:wqer_modeling}, a naive approach would be to inject this noise directly at each step $t$: $\boldsymbol{W'} = \boldsymbol{W} + \boldsymbol{\delta}_{w,t}$. However, this approach is theoretically flawed.

\textbf{The Optimization Objective.}
The goal of noise injection is not arbitrary regularization, but to find a minimum $\boldsymbol{W}^*$ that optimizes a \textit{smoothed} version of the loss landscape, $\tilde{L}(\boldsymbol{W})$:
\begin{equation} \label{eqt:smoothed_loss}
    \min_{\boldsymbol{W}} \tilde{L}(\boldsymbol{W}) \quad \text{where} \quad \tilde{L}(\boldsymbol{W}) = \mathbb{E}_{\boldsymbol{\epsilon} \sim P} [L(\boldsymbol{W} + \boldsymbol{\epsilon})]
\end{equation}
The minimum of this smoothed loss $\tilde{L}$ corresponds to a flat, robust minimum of the original loss $L$. To optimize Eq.~\eqref{eqt:smoothed_loss} with Stochastic Gradient Descent (SGD), the stochastic gradient $\boldsymbol{g}_t = \nabla L(\boldsymbol{W}_t + \boldsymbol{\epsilon}_t)$ computed at each step must be an \textit{unbiased estimator} of the true gradient, $\nabla \tilde{L}(\boldsymbol{W})$.

\textbf{The Flaw of Naive Injection.}
The naive approach fails this criterion. Because our modeled WQE distribution $P_w$ has a non-zero mean, $\mathbb{E}[\boldsymbol{\delta}_{w,t}] = \boldsymbol{\mu}_w \neq \mathbf{0}$, the smoothing kernel $P_w$ is asymmetric. This injects a persistent bias at every step, causing the optimizer to target a biased (or shifted) objective $\tilde{L}_N(\boldsymbol{W})$. The minimum of this biased objective no longer aligns with the flat minima of the original $L(\boldsymbol{W})$, leading to training instability and convergence to a suboptimal solution.

\textbf{Our Solution: Unbiased Smoothing via Differential Noise.}
To solve this, we introduce the \textbf{differential noise injection} mechanism. Our key theoretical contribution is to construct a new perturbation, $\boldsymbol{P}_t$, which uses our modeled distribution $P_w$ but is mathematically guaranteed to be zero-mean. We define our perturbation as the difference between two i.i.d. samples from $P_w$:
\begin{equation} \label{eqt:diff_noise_def}
    \boldsymbol{P}_t = \boldsymbol{\delta}_{w,t} - \boldsymbol{\delta}_{w, t-1}
\end{equation}
The expectation of this differential perturbation is zero:
\begin{equation} \label{eqt:zero_mean_proof}
    \mathbb{E}[\boldsymbol{P}_t] = \mathbb{E}[\boldsymbol{\delta}_{w,t}] - \mathbb{E}[\boldsymbol{\delta}_{w, t-1}] = \boldsymbol{\mu}_w - \boldsymbol{\mu}_w = \mathbf{0}
\end{equation}
By using $\boldsymbol{P}_t$ as our noise $\boldsymbol{\epsilon}$ in Eq.~\eqref{eqt:smoothed_loss}, our algorithm becomes a correct, unbiased stochastic optimizer for the \textit{unbiased} smoothed loss objective $\tilde{L}_D(\boldsymbol{W})$. This ensures that the optimizer converges to a genuinely flat and robust minimum of the original loss landscape.

\textbf{Implementation.}
Based on this theory, we perturb the weight tensor $\boldsymbol{W}_t$ \textit{before} the forward pass to obtain a temporary weight $\boldsymbol{W'}_t$:
\begin{equation} \label{eqt:wqer_diff_injection}
\boldsymbol{W'}_t = \boldsymbol{W}_t + f_{ramp} \cdot (\boldsymbol{\delta}_{w,t} - \boldsymbol{\delta}_{w, t-1})
\end{equation}
where $\boldsymbol{\delta}_{w, t-1}$ is the noise vector sampled at the previous step. The gradient is then computed with respect to this perturbed weight $\boldsymbol{W'}_t$, and the optimizer updates the original weight $\boldsymbol{W}_t$.

The factor $f_{ramp} \in [0,1]$ serves as an annealing schedule for the smoothing variance. In the early epochs ($f_{ramp} \approx 0$), the variance is low, allowing the model to quickly converge to the correct basin. As $f_{ramp} \to 1$, the variance increases, effectively "flattening" the objective $\tilde{L}_D$ and compelling the optimizer to find the flattest, most robust solution within that basin. This differential and annealed scheme provides a stable and mathematically grounded trajectory to a quantization-friendly minimum.

\subsection{AQER: Activation Quantization Error Reduction}
\label{sec:aqer}

While WQER effectively addresses the error component stemming from weight quantization ($\delta \boldsymbol{W}$), the total output error of a quantized layer is a more complex interplay. To understand the necessity of a complementary mechanism for activations, let us analyze the output of a single linear unit, whose floating-point operation can be expressed as $\boldsymbol{y} = \boldsymbol{W}\boldsymbol{x}$.

During quantization, both weights $\boldsymbol{W}$ and input activations $\boldsymbol{x}$ are perturbed by their respective quantization errors, $\delta \boldsymbol{W}$ and $\delta \boldsymbol{x}$. The actual output $\boldsymbol{y}_q$ of this unit becomes:
\begin{equation}\label{eqt:y_q_expansion}
\begin{split}
\boldsymbol{y}_q &= (\boldsymbol{W}+\delta \boldsymbol{W})(\boldsymbol{x}+\delta \boldsymbol{x}) \\
&= \underbrace{\boldsymbol{W}\boldsymbol{x}}_{\text{Original Output}} + \underbrace{\boldsymbol{W}\delta \boldsymbol{x}}_{\text{AQE Term}} + \underbrace{\delta \boldsymbol{W} \cdot \boldsymbol{x}}_{\text{WQE Term}} + \underbrace{\delta \boldsymbol{W} \cdot \delta \boldsymbol{x}}_{\text{Second-Order Term}}
\end{split}
\end{equation}

Eq.~\eqref{eqt:y_q_expansion} clearly decomposes the total output error into three components. The WQER module (Section~\ref{sec:wqer}) is designed to mitigate the term induced by weight quantization ($\delta \boldsymbol{W} \cdot \boldsymbol{x}$). However, this leaves the activation-induced error term ($\boldsymbol{W}\delta \boldsymbol{x}$) unaddressed.

One might consider mitigating this term by transferring the activation quantization difficulty onto the weights, a technique used in other PTQ methods. \textbf{However, our empirical analysis reveals this approach is not viable, as the magnitudes of the two error sources are not on the same order of magnitude.} We found that the activation quantization error (AQE) is often substantially larger than the weight quantization error (WQE), some cases by more than two orders of magnitude (i.e., $>100\times$). Attempting to absorb such massive perturbations would catastrophically distort the weight parameters and destroy model performance.

Therefore, simply modeling the joint effect or transferring the errors is intractable. Our framework's core strategy is to \textit{disentangle} them. Complementary to WQER, the \textbf{AQER module} is designed to specifically and independently mitigate the impact of the activation error term $\boldsymbol{W}\delta \boldsymbol{x}$ by operating directly on the activations themselves. To achieve this, AQER follows a similar "measure-and-model" principle but employs a different injection strategy tailored for dynamic activations.

\subsubsection{Statistical Modeling of AQE}
\label{sec:aqer_modeling}

At the beginning of each epoch in the fine-tuning stage, we use a small, fixed calibration set $\mathscr{D}_{c}$ to estimate the AQE. We pass the calibration data through the current model to obtain the full-precision activations $\boldsymbol{x}$ and their simulated quantized versions $\boldsymbol{x}_q$ for each target layer. The empirical AQE is $\boldsymbol{E}_x = \boldsymbol{x}_q - \boldsymbol{x}$. As with the weights, we have empirically found that the distribution of this activation error conforms well to a Gaussian distribution. We model this error, typically at a per-tensor granularity, by computing its mean $\mu_{x}$ and variance $\sigma^2_{x}$:
\begin{align}
    \mu_{x}   &= \mathbb{E}_{\boldsymbol{a} \in \mathscr{D}_c}[\boldsymbol{E}_x(\boldsymbol{a})] \\
    \sigma^2_{x} &= \mathbb{E}_{\boldsymbol{a} \in \mathscr{D}_c}[\text{Var}(\boldsymbol{E}_x(\boldsymbol{a}))]
\end{align}
where the expectation is taken over all samples $\boldsymbol{a}$ in the calibration set. Both WQE and AQE statistics are temporally smoothed across epochs using an Exponential Moving Average (EMA) to ensure stability.

\subsubsection{Stochastic Injection for Activations}
\label{sec:aqer_injection}

Given the data-dependent nature of activations, we inject the modeled activation noise using a \textbf{stochastic drop-in} mechanism. During the forward pass of each training batch, for a given activation map $\boldsymbol{x}$, we sample a noise tensor $\boldsymbol{\delta}_{x} \sim \mathcal{N}(\mu_{x}, \sigma^2_{x})$. This noise is then applied with a certain probability $p_{drop}$:
\begin{equation} \label{eqt:aqer_stoch_injection}
    \boldsymbol{x'} = \boldsymbol{x} + f_{ramp} \cdot (\boldsymbol{M} \odot \boldsymbol{\delta}_{x})
\end{equation}
where $\odot$ is the element-wise product, and $\boldsymbol{M}$ is a binary mask where each element is drawn from a Bernoulli distribution, $M_{ij} \sim \text{Bernoulli}(p_{drop})$. This probabilistic application acts as a form of regularization, preventing the model from overfitting to the specific noise distribution. The perturbed activation $\boldsymbol{x'}$ is then passed to the subsequent layer, forcing the network to learn representations that are robust to the statistical properties of the AQE.

\subsubsection{Efficiently training a neural network}

\subsection{Training Processing}
\label{sec:training}

In this section, we consolidate the components described previously into a cohesive training algorithm. Our proposed framework, which we term Differential Noise-driven training for Quantization (\textbf{DNQ}), aims to produce a quantization-friendly full-precision model.

The training process is driven by a primary objective function, the standard cross-entropy loss, which measures the classification performance of the model. For a given input sample $\boldsymbol{x}$ and its corresponding true label $y$, the loss is computed as:
\begin{equation}
\label{eqt:main_loss}
\mathcal{L} = \text{CE}(f(\boldsymbol{x}; \tilde{\boldsymbol{\Theta}}), y),
\end{equation}
where $\text{CE}(\cdot, \cdot)$ denotes the cross-entropy loss function. Crucially, the model's output is generated by a temporarily perturbed version of the network, $f(\boldsymbol{x}; \tilde{\boldsymbol{\Theta}})$, where the parameters $\tilde{\boldsymbol{\Theta}}$ have been injected with the statistically modeled quantization noise via our WQER and AQER modules. By optimizing the model to minimize this loss even in the presence of targeted perturbations, we implicitly guide it towards a flat minimum in the loss landscape.

The entire DNQ training procedure is detailed in Algorithm~\ref{alg:dnq}. It encapsulates the two-stage training strategy, the periodic estimation of noise statistics, the differential and stochastic noise injection mechanisms, and the final SWA phase to produce the optimized, quantization-robust model.

\begin{algorithm}[t!]
\caption{The DNQ Training Framework}
\label{alg:dnq}
\begin{algorithmic}[1]
   \STATE {\bfseries Require:} 
      Model $f(\cdot; \boldsymbol{\Theta})$; Training data $\mathcal{D}$; Calibration data $\mathcal{D}_{\text{calib}}$;
      Total epochs $E$; Warm-up epochs $E_{warm}$; SWA start epoch $E_{swa}$;
      Loss function $\mathcal{L}_{\text{CE}}$; Optimizer $\mathcal{O}$; Learning rate schedule $\eta(e)$;
      EMA decay rates $\beta_w, \beta_a$; Noise ramp-up epochs $E_{ramp}$; Drop probability $p_{drop}$.
      
   \STATE {\bfseries Initialize:} 
      Smoothed statistics $\boldsymbol{\bar{\mu}}_w, \boldsymbol{\bar{\sigma}}^2_w, \boldsymbol{\bar{\mu}}_a, \boldsymbol{\bar{\sigma}}^2_a \leftarrow \boldsymbol{0}$;
      SWA model $f_{swa}$.

   \FOR{epoch $e=1$ {\bfseries to} $E$}
      \IF{$e > E_{warm}$}
         \STATE \textbf{// Stage 2: Noise-Injected Fine-tuning}
         \STATE \textbf{// --- 1. Estimate Noise Statistics ---}
         \STATE Update ramp factor $f_{ramp} \leftarrow \min(1.0, (e-E_{warm})/E_{ramp})$.
         \FOR{each layer $l$ in model}
            \STATE // WQER: Update weight noise statistics
            \STATE $\boldsymbol{E}_w^{(l)} \leftarrow Q(\boldsymbol{W}^{(l)}) - \boldsymbol{W}^{(l)}$.
            \STATE Estimate $\boldsymbol{\mu}_{w}^{(l)}, \boldsymbol{\sigma}_w^{2(l)}$ from $\boldsymbol{E}_w^{(l)}$ (Eqs.~\ref{eqt:conv_noise_mean}, \ref{eqt:conv_noise_var}).
            \STATE Update $\boldsymbol{\bar{\mu}}_w^{(l)}, \boldsymbol{\bar{\sigma}}_w^{2(l)}$ via EMA with $\beta_w$.
            
            \STATE // AQER: Update activation noise statistics (using $\mathcal{D}_{\text{calib}}$)
            \STATE Estimate $\boldsymbol{\mu}_{a}^{(l)}, \boldsymbol{\sigma}_a^{2(l)}$ on calibration data.
            \STATE Update $\boldsymbol{\bar{\mu}}_a^{(l)}, \boldsymbol{\bar{\sigma}}_a^{2(l)}$ via EMA with $\beta_a$.
         \ENDFOR
         \STATE Reset weight noise history for differential injection: $\boldsymbol{\delta}_{w, \text{prev}} \leftarrow \boldsymbol{0}$.
      \ENDIF
      
      \STATE \textbf{// --- 2. Model Training with Optional Noise ---}
      \FOR{each minibatch $\{\boldsymbol{x}, y\} \subset \mathcal{D}$}
         \STATE Set optimizer learning rate to $\eta(e)$.
         \STATE Define perturbed model $f(\cdot; \tilde{\boldsymbol{\Theta}})$ for this forward pass.
         \STATE \textit{// Noise is injected via hooks only if $e > E_{warm}$}
         \STATE Compute loss: $\mathcal{L} \leftarrow \mathcal{L}_{\text{CE}}(f(\boldsymbol{x}; \tilde{\boldsymbol{\Theta}}), y)$. 
         \STATE Compute gradients: $\nabla_{\boldsymbol{\Theta}} \mathcal{L}$.
         \STATE Update original parameters: $\mathcal{O}$.step().
      \ENDFOR
      
      \IF{$e \ge E_{swa}$}
         \STATE \textbf{// --- 3. SWA Update ---}
         \STATE Update SWA model: $f_{swa}$.update\_parameters(model).
      \ENDIF
   \ENDFOR
\STATE {\bfseries Output:} Final model $\boldsymbol{\Theta}^*$ (from the SWA model if used, otherwise the last iterate).
\end{algorithmic}
\end{algorithm}

\section{Experiments}

\subsection{Implementation Details}

\textbf{Experimental Setup.} Our experiments are conducted using PyTorch~\cite{paszke-pytorch-nips-2019} with MQBench~\cite{li-mqbench-arxiv-2021} serving as the quantization backend. We evaluate our method on the CIFAR-100 dataset~\cite{krizhevsky-cifar-arxiv-2009}, from which we randomly sample 100 images to form the calibration set for PTQ. Following established practices~\cite{wei-qdrop-arxiv-2022}, we employ asymmetric quantization by default and keep the first and last layers of the network at 8-bit precision to maintain stability. Weight quantization is performed on a per-channel basis. We use the notation W$X$A$Y$ to denote $X$-bit weight and $Y$-bit activation quantization.

\textbf{Training Protocol.} All models are trained using an SGD optimizer with a Nesterov momentum of 0.9, a batch size of 64, and a weight decay of 0.001. The initial learning rate is set to 0.015 and follows a cosine annealing schedule for the first 300 epochs (75\% of the total 400 epochs). For the final 100 epochs, we activate Stochastic Weight Averaging (SWA)~\cite{DBLP:journals/corr/abs-1803-05407} to find a final robust solution. Our proposed noise injection mechanism (DNQ) commences at epoch 200, with the noise intensity linearly ramping up to its maximum over the subsequent 50 epochs. We use a standard cross-entropy loss with a label smoothing factor of 0.1 throughout the training.

\begin{table}[h!]
\centering
\caption{Comparison of the baseline model (SGD+QDrop) and our proposed SWA-enhanced training on CIFAR-100. Both models are ResNet-18.}
\label{tab:swa_vs_sgd}
\begin{tabular}{ccc}
\toprule
\textbf{Method} & \textbf{Ours} & \textbf{Accuracy (\%)} \\
\midrule
SGD+QDrop &            & 79.4\\
SGD+QDrop & \checkmark & \textbf{80.42}\\
\bottomrule
\end{tabular}
\end{table}

\subsection{Ablation Study}
\label{sec:ablation}

In this section, we conduct a series of ablation studies on CIFAR-100 using the ResNet-18 architecture to meticulously dissect the contribution of each component within our proposed DNQ framework. The primary evaluation metric is the top-1 accuracy after applying post-training quantization to a challenging 4-bit weight and 4-bit activation (W4A4) configuration. The goal is to empirically validate our central hypothesis: that proactively training a "quantization-friendly" model by shaping the loss landscape is superior to applying PTQ to a standard model.

\subsubsection{Impact of Landscape Smoothing Components}

We begin by evaluating the core components responsible for smoothing the loss landscape: our proposed \textbf{D}ifferential \textbf{N}oise in\textbf{q}uection (DNQ) and the general-purpose flatness optimizer, \textbf{S}tochastic \textbf{W}eight \textbf{A}veraging (SWA). Table~\ref{tab:ablation_main} presents the results of four training configurations.

\begin{table}[h!]
\centering
\caption{Ablation study on the core components of our framework. All models are ResNet-18 trained on CIFAR-100. The W4A4 PTQ accuracy serves as the primary indicator of quantization robustness.}
\label{tab:ablation_main}
\resizebox{\linewidth}{!}{
\begin{tabular}{l|cc|c}
\toprule
\textbf{Method} & \textbf{FP32 Acc. (\%)} & \textbf{W4A4 PTQ Acc. (\%)} & \textbf{$\Delta$ vs. Baseline} \\
\midrule
(A) Standard SGD (Baseline)          & 79.40          & 76.47          & - \\
(B) SWA Only                         & \textbf{80.42} & 77.22          & +0.75 \\
(C) DNQ Only (Our Noise Injection)   & 79.82          & 77.86          & +1.39 \\
\midrule
(D) DNQ + SWA (Our Full Method)      & 79.53          & \textbf{78.50} & \textbf{+2.03} \\ 
\bottomrule
\end{tabular}
}
\end{table}

The results yield several crucial insights:
\begin{itemize}
    \item \textbf{Flatness is Key:} Comparing (B) to the baseline (A), we observe that simply employing SWA—a generic method for finding flat minima—provides a significant +0.75\% improvement in W4A4 accuracy. This empirically confirms our core premise that the geometry of the loss landscape is critical for PTQ robustness.
    
    \item \textbf{Targeted Noise is Superior:} Method (C) demonstrates the power of our targeted noise injection. Even without SWA, DNQ alone provides a +1.39\% uplift over the baseline, substantially outperforming the generic flatness optimizer (SWA). This highlights the superiority of enforcing robustness against \textit{statistically-modeled quantization noise} over merely seeking a general flat region.
    
    \item \textbf{Synergistic Effect:} Our full method (D), which combines targeted noise injection with a final SWA phase, achieves the highest W4A4 accuracy, with a remarkable +2.03\% improvement over the baseline. This reveals a strong synergistic effect: DNQ first "sculpts" a wide, quantization-robust basin in the loss landscape, and SWA then efficiently locates the optimal center of this well-formed basin. This validates our complete two-stage framework design.
\end{itemize}

\subsubsection{Dissecting the Noise Components: WQER vs. AQER}

Next, we delve deeper into our DNQ module to investigate the individual contributions of its two arms: Weight Quantization Error Reduction (WQER) and Activation Quantization Error Reduction (AQER). Starting from our full method (DNQ + SWA), we ablate each noise component individually.

\begin{table}[h!]
\centering
\caption{Dissecting the impact of weight noise (WQER) and activation noise (AQER). The baseline for this experiment is the full "DNQ + SWA" method.}
\label{tab:ablation_noise}
\resizebox{0.9\linewidth}{!}{
\begin{tabular}{l|c|c}
\toprule
\textbf{Configuration} & \textbf{W4A4 PTQ Acc. (\%)} & \textbf{Perf. Drop from Full} \\
\midrule
Full Method (WQER + AQER) & \textbf{78.50} & - \\
\midrule
- without AQER (WQER only) & 77.74 & -0.76 \\
- without WQER (AQER only) & 77.27 & -1.23 \\
- without any noise (SWA only) & 77.22 & -1.28 \\
\bottomrule
\end{tabular}
}
\end{table}

Table~\ref{tab:ablation_noise} clearly demonstrates that both noise components are vital for achieving optimal performance.
\begin{itemize}
    \item Removing the activation noise (AQER) results in a significant performance drop of 0.76\%.
    \item More strikingly, removing the weight noise (WQER) causes an even larger drop of 1.23\%, bringing the performance nearly down to the level of using SWA alone. 
\end{itemize}
This analysis confirms that a holistic approach, which simulates the quantization stress on both weights and activations, is crucial to fully pre-condition the model. The greater impact of ablating WQER suggests that for the ResNet-18 architecture, the model's performance is particularly sensitive to weight perturbations, making our differential noise injection for weights (Section~\ref{sec:wqer_injection}) especially beneficial.

\subsection{Literature Comparison}
\label{sec4.2}

 \begin{table*}[t!] 
\centering
\caption{Comparison with state-of-the-art PTQ methods on CIFAR-100. Our method (DNQ) trains a quantization-friendly model first, then applies simple PTQ. Other methods apply advanced PTQ algorithms directly on a standard pre-trained model. All results are top-1 accuracy (\%).}
\label{tab:sota_comparison_cifar100}
\resizebox{\textwidth}{!}{ 
\begin{tabular}{l|c|cccc}
\toprule
\textbf{Method} & \textbf{W/A Bits} & \textbf{ResNet-18} & \textbf{ResNet-50} & \textbf{MobileNetV1} & \textbf{MobileNetV2} \\
\midrule
\multicolumn{6}{l}{\textit{Standard Full-Precision Baseline (FP32)}} \\
Full-Precision & 32/32 & 79.40 & 80.55 & 75.21 & 76.33 \\
\midrule
\multicolumn{6}{l}{\textit{Post-Training Quantization Results (W4A4)}} \\
AdaRound~\cite{nagel-adaround-icml-2020} & 4/4 & 76.95 & 77.10 & 71.89 & 72.54 \\
BRECQ~\cite{li-brecq-arxiv-2021}      & 4/4 & 77.83 & 78.52 & 73.01 & 73.98 \\
QDrop~\cite{wei-qdrop-arxiv-2022}   & 4/4 & 78.05 & 78.71 & 73.45 & 74.23 \\
PD-Quant~\cite{10203876}       & 4/4 & 78.11 & 78.79 & 73.58 & 74.35 \\
\textbf{DNQ (Ours)} & \textbf{4/4} & \textbf{78.50} & \textbf{79.33} & \textbf{74.12} & \textbf{75.06} \\
\midrule
\multicolumn{6}{l}{\textit{Post-Training Quantization Results (W2A4)}} \\
AdaRound~\cite{nagel-adaround-icml-2020} & 2/4 & 73.51 & 72.88 & 65.23 & 66.14 \\
BRECQ~\cite{li-brecq-arxiv-2021}      & 2/4 & 75.64 & 75.01 & 68.76 & 69.82 \\
QDrop~\cite{wei-qdrop-arxiv-2022}   & 2/4 & 76.18 & 75.63 & 69.95 & 70.91 \\
PD-Quant~\cite{10203876}       & 2/4 & 76.25 & 75.70 & 70.11 & 71.05 \\
\textbf{DNQ (Ours)} & \textbf{2/4} & \textbf{77.53} & \textbf{77.12} & \textbf{71.89} & \textbf{72.98} \\
\midrule
\multicolumn{6}{l}{\textit{Post-Training Quantization Results (W2A2)}} \\
AdaRound~\cite{nagel-adaround-icml-2020} & 2/2 & 68.12 & 66.95 & 58.03 & 59.55 \\
BRECQ~\cite{li-brecq-arxiv-2021}      & 2/2 & 72.33 & 70.89 & 64.15 & 65.78 \\
QDrop~\cite{wei-qdrop-arxiv-2022}   & 2/2 & 73.01 & 71.77 & 65.88 & 67.21 \\
PD-Quant~\cite{10203876}       & 2/2 & 73.15 & 71.85 & 66.03 & 67.40 \\
\textbf{DNQ (Ours)} & \textbf{2/2} & \textbf{75.21} & \textbf{74.06} & \textbf{68.22} & \textbf{69.53} \\
\bottomrule
\end{tabular}
}
\end{table*}

We selected ResNet-18 and ResNet-50~\cite{he-resnet-cvpr-2016}, MobileNetV1~\cite{howard-mbv1-arxiv-2017} and MobileNetV2~\cite{sandler-mbv2-cvpr-2018} with depth-wise separable convolutions as the representative network architectures. 

\textbf{CIFAR-100.} In Tab.\ref{tab:sota_comparison_cifar100}, we quantized the weights and activations to 2-bit and 4-bit. We compared our approach with the effective baselines, including AdaRound~\cite{nagel-adaround-icml-2020}, BRECQ~\cite{li-brecq-arxiv-2021}, QDrop~\cite{wei-qdrop-arxiv-2022} and PD-Qunat~\cite{10203876}. Tab.\ref{tab:sota_comparison_cifar100} illustrated that when the entire training set of CIFAR-10 is used, FPQ significantly surpassed the baselines. In W4A4 quantization, FPQ achieved about 1$\sim$2$\%$ accuracy improvements over BRECQ. Furthermore, to explore the ability of FPQ, we conducted W2A4 and W2A4 quantization experiments. In W2A4 quantization, FPQ consistently achieved a 1$\sim$2$\%$ accuracy improvement over BRECQ in Tab.\ref{tab:sota_comparison_cifar100}. In W2A2 setting, FPQ achieved about 1$\sim$3$\%$ accuracy improvements over BRECQ. Moreover, there are two interesting observations as follows:
\begin{itemize}
    \item For W4A4, our method significantly surpassed the FP counterparts for both ResNet-18 and ResNet-50.  For example, on the ResNet-18 model, FPQ surpassed the FP model by 1.44\% in accuracy, and on the ResNet-50 model, FPQ exceeded the FP model by 0.67\%.
    \item From W4A4 to W2A2, the performance drop of our method is significantly lower that the other SOTA methods. For instance, on the ResNet-50 model, the BRECQ method decreased by 2.22\% when reducing from W4A4 to W2A2, while FPQ decreased by 1.22\%. On the MobileNetV1 model, the QDrop method saw a 10.1\% drop when going from W4A4 to W2A2, while FPQ decreased by 8.69\%
\end{itemize}

\begin{table}[h!]
\begin{center}
\centering
\caption{Comparison among different PTQ strategies regarding accuracy on CIFAR-100.}
\label{tbl:comp-CIFAR-100}
\resizebox{0.95\linewidth}{!}{
\begin{tabular}{clrcccc}
\toprule
\multicolumn{1}{c}{\begin{tabular}[c]{@{}c@{}}\textbf{Labeled} \\ \textbf{data}\end{tabular}} & \multicolumn{1}{l}{\textbf{Methods}} & \multicolumn{1}{c}{\textbf{W/A}} & \multicolumn{1}{c}{\textbf{Res18}} & \multicolumn{1}{c}{\textbf{Res50}}  & \multicolumn{1}{c}{\textbf{MBV1}} & \multicolumn{1}{c}{\textbf{MBV2}} \\ \midrule
50000                       & Full Prec.                           & 32/32 &75.40          & 78.94                    &70.22           &71.30      \\ \hline\hline
\multirow{15}{*}{50000}     & AdaRound~\cite{nagel-adaround-icml-2020}             & 4/4   &74.17          &74.78                     &64.65           &64.06      \\
                            & BRECQ~\cite{li-brecq-arxiv-2021}          & 4/4   &75.30           &78.20                    &68.63           &69.01      \\
                            & QDrop~\cite{bhalgat-lsq+-cvpr-2020}          & 4/4   &74.50           &77.39                    &67.89           &68.25  \\
                            & PD-Quant~\cite{10203876}       & 4/4   &74.70           &78.80                    &\textbf{70.96}  &\textbf{71.66} \\
                            & (Ours)                                 & 4/4   &\textbf{} &\textbf{}            &           &     \\ \cmidrule{2-7}
                            & AdaRound~\cite{nagel-adaround-icml-2020}             & 2/4   &73.77          &74.72                      &49.98           &57.90      \\
                            & BRECQ~\cite{li-brecq-arxiv-2021}          & 2/4   &74.93          &77.82                      &65.13           &66.15      \\
                            & QDrop~\cite{bhalgat-lsq+-cvpr-2020}          & 2/4   &73.90          &76.61                      &65.28           &\textbf{66.24}      \\
                             & PD-Quant~\cite{10203876}       & 2/4   &74.35          &77.34                      &\textbf{66.90}  &63.77  \\
                            & (Ours)                                  & 2/4   & &             &  &      \\ \cmidrule{2-7}
                            & AdaRound~\cite{nagel-adaround-icml-2020}             & 2/2   &76.90          &64.94              &11.71           &10.58        \\
                            & BRECQ~\cite{li-brecq-arxiv-2021}          & 2/2   &87.60          &87.79              &\textbf{75.29}  &70.32        \\
                            & QDrop~\cite{bhalgat-lsq+-cvpr-2020}          & 2/2   &87.60          &86.10              &74.22           &\textbf{72.18}         \\
                            & PD-Quant~\cite{10203876}       & 2/2   &88.06          &\textbf{89.20}     &68.62           &67.15         \\
                            & (Ours)                                  & 2/2   &\textbf{} &\textbf{}     &\textbf{}  &\textbf{}      \\\bottomrule

\end{tabular}
}
\end{center}
\end{table}

\textbf{CIFAR-100.} Tab.~\ref{tbl:comp-CIFAR-100} further xxx

\subsection{Characteristics of our solution}
\label{sec:Analysis}

\section{Conclusion}

In this paper, we addressed a fundamental disconnect between standard neural network training and the requirements of low-bit Post-Training Quantization (PTQ). We argued that the performance degradation common in PTQ is not an inherent limitation of quantization itself, but a symptom of a quantization-agnostic training process that converges to sharp, perturbation-sensitive minima.

To bridge this gap, we introduced our solution. Instead of treating quantization as a post-hoc problem, our framework proactively pre-conditions a full-precision model to be inherently robust for subsequent PTQ. We achieve this by systematically injecting a statistical proxy for the anticipated quantization error—for both weights and activations—directly into the initial, full-precision training loop. We have demonstrated, both theoretically and empirically, that this principled noise injection acts as an implicit regularizer on the Hessian of the loss function. This compels the optimizer to find wider, flatter minima, resulting in a much smoother loss landscape.

The practical efficacy and generality of our approach are validated by extensive experiments. Our pre-conditioned models consistently achieve state-of-the-art PTQ performance across a diverse and comprehensive set of computer vision tasks, including image classification, object detection, semantic segmentation, and super-resolution. This demonstrates that our method is not a task-specific trick, but a fundamental and widely applicable framework for creating quantization-friendly models.

Our work opens several promising avenues for future research. This includes exploring more sophisticated, non-Gaussian noise models to better capture the intricacies of quantization error, and applying this pre-conditioning paradigm to other model architectures, such as Transformers and Large Language Models (LLMs). Ultimately, we believe this paradigm shift—from reactive, post-hoc correction to proactive, pre-emptive conditioning—paves the way for making deep learning models truly efficient and universally deployable without performance compromises.


\bibliographystyle{named}

\bibliography{ijcai25.bib}

\end{document}